\documentclass{article}

\usepackage{microtype}
\usepackage{graphicx}
\usepackage{subfigure}
\usepackage{booktabs} 
\graphicspath{{img/}}
\usepackage{multirow}
\usepackage{siunitx}

\usepackage{hyperref}

\usepackage[accepted]{icml2025}

\usepackage{amsmath}
\usepackage{amssymb}
\usepackage{mathtools}
\usepackage{amsthm}
\usepackage{verbatim}
\usepackage{tcolorbox}  
\tcbuselibrary{breakable}

\usepackage{color}
\definecolor{keywordcolor}{rgb}{0.7, 0.1, 0.1}   
\definecolor{commentcolor}{rgb}{0.4, 0.4, 0.4}   
\definecolor{symbolcolor}{rgb}{0.0, 0.1, 0.6}    
\definecolor{sortcolor}{rgb}{0.1, 0.5, 0.1}      
\definecolor{errorcolor}{rgb}{1, 0, 0}           
\definecolor{stringcolor}{rgb}{0.5, 0.3, 0.2}    
\usepackage{listings}

\usepackage[capitalize,noabbrev]{cleveref}

\theoremstyle{plain}

\theoremstyle{definition}

\theoremstyle{remark}

\usepackage[textsize=tiny]{todonotes}

\icmltitlerunning{FMC: Formalization of Natural Language Mathematical Competition Problems}

\newcommand{\dataset}{\textit{FMC}}

\begin{document}

\twocolumn[
\icmltitle{FMC: Formalization of Natural Language Mathematical Competition Problems}



\icmlsetsymbol{equal}{*}

\begin{icmlauthorlist}
\icmlauthor{Jiaxuan Xie}{sch1}
\icmlauthor{Chengwu Liu}{sch1}
\icmlauthor{Ye Yuan}{sch1}
\icmlauthor{Siqi Li}{sch1}
\icmlauthor{Zhiping Xiao}{sch2}
\icmlauthor{Ming Zhang}{sch1}
\end{icmlauthorlist}

\icmlaffiliation{sch1}{State Key Laboratory for Multimedia Information Processing, School of Computer Science, PKU-Anker LLM Lab, Peking University, Beijing, China}
\icmlaffiliation{sch2}{Paul G. Allen School of Computer Science and Engineering, University of Washington, U.S.A.}

\icmlcorrespondingauthor{Zhiping Xiao}{patxiao@uw.edu}
\icmlcorrespondingauthor{Ming Zhang}{mzhang\_cs@pku.edu.cn}

\icmlkeywords{Machine Learning, ICML}

\vskip 0.3in
]



\printAffiliationsAndNotice{The source code and dataset are available in \url{https://github.com/JadeXie1205/FMC}.}

\begin{abstract}
Efficient and accurate autoformalization methods, which leverages large-scale datasets of extensive natural language mathematical problems to construct formal language datasets, are key to advancing formal mathematical reasoning. In this paper, we propose an autoformalization pipeline based on large language models with error feedback, achieving a fully automatic and training-free formalization approach. Using this pipeline, we curate an Olympiad-level dataset aligning natural language problems with Lean formalizations. The dataset comprises $3,922$ mathematical problems in natural language and $9,787$ in Lean, of which $64.46\%$ were assessed as at least above-average quality, making it suitable as a benchmark for automated theorem provers. Additionally, we investigate the formalization and reasoning capabilities of various LLMs and empirically demonstrate that few-shot learning, error feedback, and increasing sampling numbers enhance the autoformalization process.
Experiments of three automated theorem provers on the \dataset\ dataset also highlights its challenging nature and its value as a benchmark for formal reasoning tasks.
\end{abstract}

\section{Introduction}
\label{submission}
Large language models (LLMs), due to their strong textual reasoning capabilities, have been widely applied to mathematical problem reasoning. Initially developed for reasoning within natural language, LLMs face challenges such as the scarcity of complex mathematical data and the occurrence of hallucinations. To address these issues, formal languages have been introduced into LLM mathematical reasoning.

A formal language is a logical system in which statements and derivations can be verified through an interactive theorem prover, thereby mitigating the hallucination problem. However, formal reasoning introduces a new challenge---an even greater scarcity of data. To address this challenge, research on formal mathematical reasoning has primarily followed two directions: \emph{automated theorem proving} and \emph{autoformalization}, with the latter often serving as a source of training data for the former. This work focuses on the autoformalization of mathematical problems presented in natural language.

We propose an enhanced autoformalization pipeline with error feedback. Built upon the stages of \emph{formal translation -- formal verification -- back translation -- consistency check}, our pipeline collects error information from the verification and consistency stages. These errors, along with the original natural language problem, are fed back into the formalization model for a second translation, enabling the model to perform self-correction.

Using this pipeline, we construct a dataset of aligned natural language--Lean pairs, focusing specifically on mathematical problems of Olympiad difficulty. The original natural language problems are sourced from \texttt{IMOmath}, which curates problems from various national and international Olympiad competitions. After preprocessing, the problems are passed through our formalization pipeline, yielding a dataset of $3,922$ natural language problems aligned with $9,787$ formal statements.

In evaluating the pipeline, we adopt \emph{syntactic validity} and \emph{semantic consistency} as key metrics. Our method achieves a syntactic validity of $93.39\%$ and semantic consistency of $81.74\%$. Compared to a closely related recent work \texttt{Lean Workbook} \cite{41}, our dataset is based on significantly more challenging problems and outperforms it's autoformalization pipeline, which achieved $62.5\%$ syntactic validity and $17.5\%$ semantic consistency. Furthermore, we evaluate each example using a large language model across five criteria: \emph{relevance to current research}, \emph{complexity and depth}, \emph{interdisciplinary potential}, \emph{community needs and gaps}, and \emph{novelty}. Overall, $64.46\%$ of the examples are rated as above-average or above.

\paragraph{The contributions are summarized as follows:}
\begin{enumerate}
    \item We propose an enhanced autoformalization pipeline using LLMs with error-feedback, enabling a fully automated, training-free formalization process.
    \item Using this pipeline, we curate a dataset of $3,922$ aligned natural language--Lean pairs, with problems sourced from national and international mathematics Olympiads. The semantic consistency of the formalization pipeline reaches $81.74\%$. Our quality assessment reveals that $64.46\%$ of the examples are rated as above-average or higher quality, indicating the high quality of our dataset.
    \item We investigate the formalization and semantic alignment checking capabilities of different general-purpose LLMs and find that \texttt{DeepSeek-R1} remains at the forefront. Experimental results further demonstrate that few-shot learning, error feedback, and increased sampling per problem enhance autoformalization performance.
\end{enumerate}

\section{Related Work}
\subsection{Large Language Models}
Large Language Models (LLMs) represent a significant paradigm shift in the evolution of natural language processing. Typically built upon the Transformer architecture and equipped with tens or hundreds of billions of parameters, LLMs are trained on massive textual corpora. Representative models include PaLM\cite{9}, GPT-4\cite{8}, DeepSeek-V3\cite{7}, and Claude 3.7\cite{10}. Their unprecedented scale in both model size and training data enables capabilities distinct from smaller models—capabilities often referred to as emergent abilities\cite{6} These include in-context learning, instruction following, and step-by-step reasoning\cite{4}, allowing LLMs to handle complex tasks across diverse domains, including reasoning tasks.

Among various reasoning tasks, mathematical reasoning has attracted particular attention. Previous research \cite{11} has shown that using LLMs for mathematical reasoning generally involves three stages: pretraining, fine-tuning with structured data, and invoking external tools. However, reasoning within the scope of natural language using LLMs still faces significant challenges, such as the scarcity of high-difficulty data and the problem of hallucination. First, effective training of LLMs relies on large-scale, high-quality corpora. Yet for domains like university-level mathematics, mathematical competition problems, or research-level mathematical reasoning, it is difficult to obtain sufficient data that include detailed reasoning processes. A more critical issue is that LLMs are prone to \emph{hallucination} when performing mathematical reasoning---that is, they often generate plausible but logically flawed proofs. These subtle errors are hard to detect, significantly increasing the difficulty of evaluating the correctness of mathematical reasoning. Some recent efforts have incorporated self-verification mechanisms to mitigate hallucinations \cite{29}, but the results remain unsatisfactory.

As a result, some researchers have turned to formal languages for mathematical reasoning. Although data for formal languages are even more limited, their verifiability ensures the correctness and reliability of reasoning. Formal mathematical reasoning has thus become a research focus within the broader field of AI reasoning.

\subsection{Formal Languages}
Formal languages express mathematics within formal systems. They impose strict syntactic rules, and operations such as verification must adhere to logically sound inference rules\cite{11}.

Currently, formal math languages such as Isabelle (1986)~\cite{30}, Coq (1989)~\cite{31}, and Lean (2015)~\cite{18} have attracted significant attention from researchers. This study uses Lean~\cite{19}, a modern open-source theorem prover developed by Microsoft Research and CMU. Lean combines interactive and automated theorem proving, and supports reasoning in both mathematics and complex systems in computer engineering.

\texttt{Mathlib} is commonly used in formalizing mathematical theorems in Lean. It is a community-driven project aimed at building a unified library of formalized mathematics for the Lean prover. \texttt{Mathlib4}, the updated version for Lean 4, includes many important mathematical objects, pre-formalized theorems, and automation strategies. Some of its metaprograms enable non-trivial proof automation.

\subsection{Automformalization}
Autoformalization is a key research direction in formalized mathematical reasoning. It is essentially a translation task: converting natural language mathematical problems into formal theorems or proofs using computational methods. The former can serve as benchmarks for automated theorem provers, while the latter can be used as training data. Early studies\cite{32} indicate that general-purpose LLMs possess a degree of formalization ability, and training with formal proof data significantly improves the performance of automated theorem provers.

Autoformalization is the central focus of this study. More precisely, this work investigates how to construct a high-quality, aligned dataset of natural language and formal language pairs via autoformalization, to serve as a benchmark for evaluating theorem provers.

Constructing a Lean-based dataset via autoformalization involves two primary steps: sourcing the original data and conducting formalization. Data sources can be broadly classified into three categories: manual curation, natural language datasets, and automatic synthesis. Manual curation refers to the direct authoring of Lean theorems and proofs by mathematical experts. Most of the pre-proved theorems in Mathlib fall into this category. Natural language datasets such as the MATH dataset\cite{22}, GSM8K\cite{23}, and AQuA-RAT\cite{24} contain mathematical problems in natural language, which can be formalized into formal language datasets. Automatic synthesis involves the computational generation of new mathematical problems based on existing concepts and theorems. Representative works include MUSTARD\cite{33}, which first synthesizes problems in natural language before formalizing them, and STP\_Lean\cite{34}, which directly synthesizes problems in formal language, among others \cite{55, 56}.
As for formalization methods, they can be roughly divided into two types: manual annotation, as demonstrated by works such as PutnamBench\cite{35}, miniF2F\cite{36}, and ProofNet\cite{37}; and autoformalization, which is now predominantly powered by large language models.
Our analysis of existing datasets reveals the following key observations: (1) Datasets containing competition-level problems are usually manually annotated and remain relatively small in scale. (2) Extracting natural language problems from the web and automatically formalizing them is still the dominant dataset construction strategy. However, the difficulty levels of these problems frequently exhibit considerable variability. (3) Data synthesis allows for large-scale dataset generation, but remains underexplored.

This study aims to construct high-quality datasets with minimal training costs. Manual annotation is labor-intensive and prohibitively expensive, while typical data synthesis requires substantial computational resources. Therefore, we adopt an approach based on autoformalization of natural language mathematical problems.

Our autoformalization pipeline is inspired by the design of the Lean Workbook pipeline, with improvements to the base translation model and the incorporation of error feedback to enhance automation. For data selection, we focus on Olympaid-level mathematical problems to ensure sufficient difficulty and quality in the resulting dataset.

\section{Autoformalization Pipeline Design}

\begin{figure}[ht]
\vskip 0.2in
\begin{center}
\centerline{\includegraphics[width=\columnwidth]{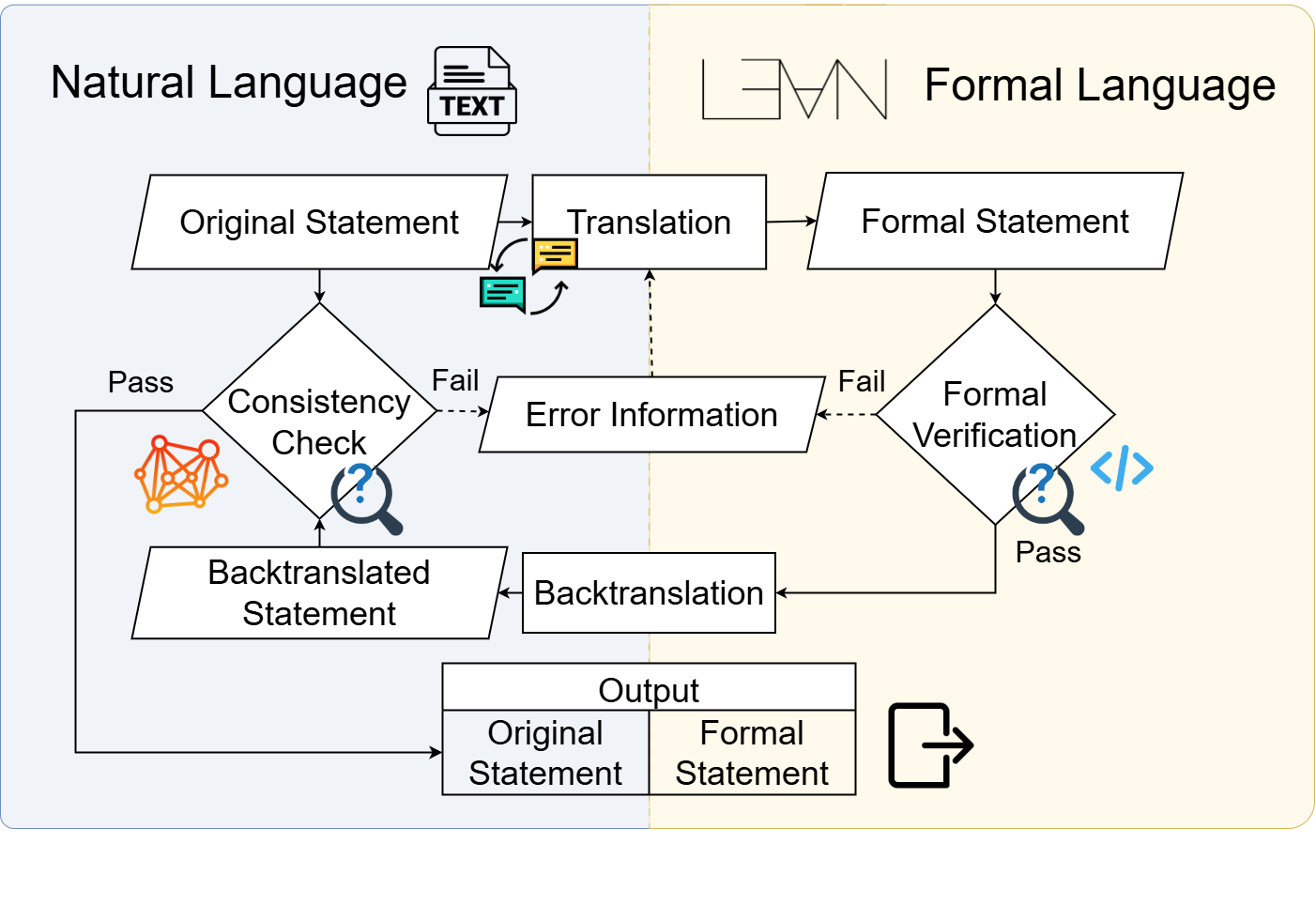}}
\caption{Autoformalization Pipeline.}
\label{fig:flow_chart}
\end{center}
\vskip -0.2in
\end{figure}

Inspired by the formalization pipeline of the Lean Workbook\cite{41}, this paper proposes an autoformalization pipeline that translates mathematical problems from natural language into Lean language with error feedback. The entire pipeline is shown in Figure~\ref{fig:flow_chart}. Each natural language mathematical problem is first translated into Lean language using a few-shot prompting approach and then formally verified by Lean REPL, i.e., Lean's syntax check. Those that pass the formal verification are translated back into natural language and compared with the original problem to perform semantic consistency checks. Statements that pass both the formal verification and semantic consistency check are regarded as successfully formalized. In cases where a statement fails either check, the corresponding error information is incorporated into a revised prompt and fed back to the translation model, enabling iterative prompt refinement driven by error feedback.

\subsection{Translation}
Within this autoformalization pipeline, the model is required to translate natural language mathematical problems into formal language in two distinct scenarios: the first involves direct translation guided by few-shot prompting, while the second leverages error feedback incorporated into the prompt following a failed attempt.

The first case occurs during the initial translation of a mathematical problem and is based on few-shot learning. In large language models, few-shot learning refers to solving tasks using prompts that include a task description and a few examples, without gradient updates or fine-tuning—this is known as in-context learning.
Experiments on GPT-3\cite{43} and related work have shown that few-shot learning can significantly improve performance on tasks involving structured formatting or translation. In this study, each translation prompt includes two fixed examples—one from algebra and one from number theory—both correctly aligned in natural language and Lean. Since problems in these two fields are common in math Olympiad contests, using such examples helps improve translation accuracy.

Additionally, since large language models generate each token by first computing a probability distribution and then sampling from it, the output inherently contains stochasticity. As a result, repeated formalization requests for the same input may yield different outputs. The temperature parameter T controls the sharpness of this distribution, thereby influencing the creativity of the generated text. 
Since formalization tasks require syntactic correctness while still benefiting from some diversity to improve translation success, this pipeline sets the temperature to 1.0 during formalization and samples each input five times.

The second case arises when a theorem fails either formal verification or semantic consistency check. In such instances, the associated error information is incorporated into the prompt and fed back to the translation model. The model then attempts to retranslate the theorem, leveraging the in-context learning capabilities of LLMs to produce a valid formal representation. Experimental results demonstrate that incorporating error feedback improves formalization accuracy, as detailed in the experimental section.

Additionally, since Lean focuses on theorem proving and cannot resolve open problems lacking explicit solutions within the statements, it is necessary to address missing solutions and proofs. For absent solutions, we depend on the model’s reasoning capabilities, expecting it to generate them during the formalization process. For missing proofs, the placeholder ":= by sorry" is employed, enabling Lean to detect the omission and signal the missing proof without triggering errors.

After comparing several frontier LLMs, this work directly employs the Deepseek-R1 \cite{44} as the translation model. As one of the state-of-the-art large language models, Deepseek-R1 possesses strong reasoning and in-context learning abilities, demonstrating promising performance in formalization. The comparison of formalization capabilities of different language models are presented in the experimental section of this paper.

\subsection{Formal Verification}

Each translated formal theorem must undergo formal verification to ensure its syntactic correctness. In this study, the formal verifier from DeepSeek-Prover\cite{51} is used to submit the Lean statement to Lean 4 REPL for validation and parsing the returned results. Specifically, formal verification is performed by invoking Lean 4’s interactive Read-Eval-Print Loop (REPL) via a Python subprocess. The formal theorem, including Mathlib import statements, is passed through standard input in JSON format, while the output records any compilation or type errors. This approach confirms the syntactic correctness of the formal theorem and provides error messages for failed compilations, facilitating iterative prompt refinement based on error feedback.

\subsection{Backtranslation}

Even when statements formalized by LLMs pass Lean’s syntax checks, they sometimes differ semantically from the original theorems—such as missing conditions, incorrect assumptions, or erroneous goals. These discrepancies clearly indicate incorrect formalization. Therefore, semantic verification is necessary to ensure that the formalized theorems correspond to the same mathematical problems described in the original statements.

Although LLMs have limited formalization capabilities, they are good at back translation—translating Lean-formalized mathematical problems back into natural language. Comparing the consistency between two natural language problem descriptions is a comparatively simple task. Therefore, this autoformalization pipeline employs a “back translation—consistency check” approach to ensure semantic equivalence. This stepwise process resembles human reasoning and, despite introducing some noise, effectively improves the overall accuracy of semantic verification. Similarly, this automated pipeline employs Deepseek-R1\cite{44} as the back translation model. 

\subsection{Consistency Check}
In consistency check, the goal is to verify whether the backtranslated natural language theorem is mathematically consistent with the original theorem. The prompt sent to the model explicitly emphasizes examining whether the two problems share the same mathematical essence, aiming to prevent incorrect judgments caused by differences in wording or contextual information. Additionally, since some formalized theorems may superficially include all elements of the original theorem but have confused conditions and goals, the prompt clearly instructs the model to verify the consistency of both conditions and goals. To facilitate iterative prompt refinement based on error feedback, the model is also required to provide a brief rationale for its judgment, enabling the use of error cases in subsequent iterations.

Similarly, this automated pipeline employs Deepseek-R1\cite{44} as the consistency checking model. Notably, due to Deepseek-R1’s strong reasoning capabilities, it often produces extensive analytical output. To facilitate the extraction of relevant information, the prompt explicitly restricts the output format.

\section{Dataset Construction}

\subsection{Data Collection}
To enhance dataset difficulty and ensure quality, all mathematical problems in this dataset are all official Olympiad problems on the \texttt{IMOmath} website\cite{26}. It covers 11 international competitions including the International Mathematical Olympiad (IMO), as well as 42 national and regional Olympiad contests. The dataset spans the years 1959 to 2011 and covers six continents: Europe, Asia, North America, South America, Oceania, and Africa. These contests are either the IMO and its regional selection rounds or other major competitions of comparable difficulty. By selecting these sources as the original natural language data, the dataset guarantees the integrity, correctness, and challenge of the mathematical theorems. The resulting aligned natural language–Lean dataset will facilitate the evaluation of automated theorem provers’ capabilities.

To collect problems, this study employed web crawling techniques to download PDF files from relevant websites, and used Optical Character Recognition (OCR) to extract problem statements from the PDFs. OCR is a technology that converts printed text and images into machine-editable formats~\cite{27}. In this work, we used Mathpix~\cite{29} as the OCR tool to convert PDF content into markdown files. Mathpix specializes in OCR, with support for multiple languages and various export formats. More importantly, it is optimized for the structured recognition of mathematical expressions, such as fractions, subscripts, superscripts, and summations, and accurately converts them into well-formed \LaTeX{} code. This capability is particularly critical for extracting mathematical problems in our study. Finally, we used regular expressions to extract the problem texts from the markdown files and organized them into JSON format. After preprocessing, we obtained a total of 6{,}980 natural language mathematical problems.

\subsection{Data Preprocessing}
To further improve the quality of the formalized data, this study conducted additional filtering on the extracted $6,980$ natural language mathematical problems. Preliminary experiments revealed that LLMs encounter significant difficulties when formalizing geometry problems into Lean.

\begin{figure}[ht]
\vspace{0.2cm}
\begin{center}
\centerline{\includegraphics[width=\columnwidth]{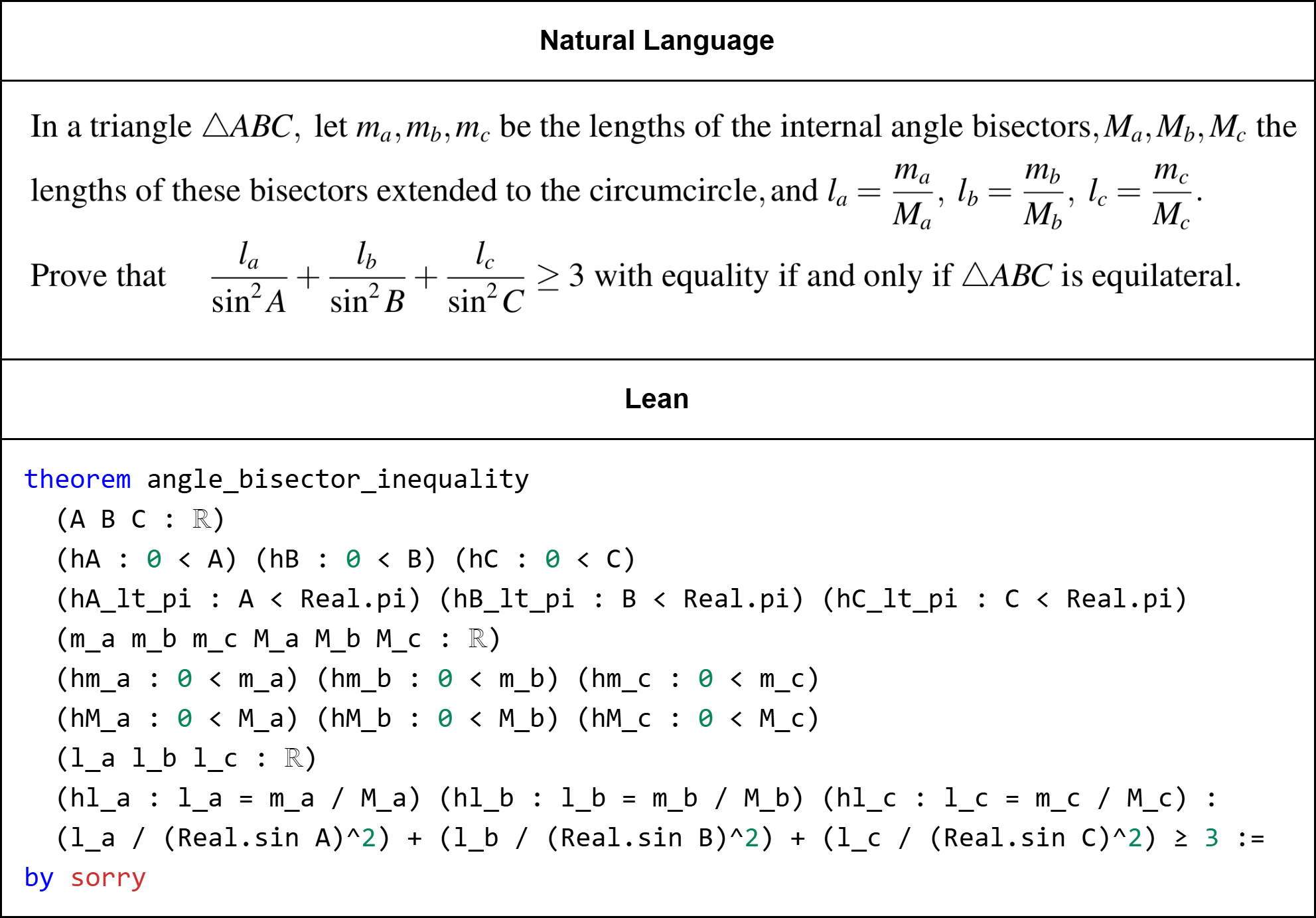}}
\vspace{-0.4cm}
\caption{A failed formalization example in geometry.}
\label{fig:geo_example}
\end{center}
\vspace{-0.9cm}
\end{figure}

Figure~\ref{fig:geo_example} presents an incorrect formalization example. The original problem, described in natural language, involves a mathematical proposition about the internal angle bisectors of a triangle, where \( m_a, m_b, m_c \) denote the lengths of the three angle bisectors, and \( M_a, M_b, M_c \) represent their extended lengths intersecting the circumcircle. The objective is to prove an inequality along with the conditions for equality. In the Lean formulation, although the angles \(\angle A\) , \(\angle B\) and \(\angle C\) are constrained between \(0\) and \(\pi\), and all six segment lengths are specified to be positive, with accurate quantitative relationships given for \( l_a, l_b, l_c \) relative to the six known segments and the final goal stated, the essential triangle constraint is missing. For example, the condition that the interior angles of \( \triangle ABC \) sum to 180° is absent, and constraints involving angle bisectors, the circumcircle, and their intersections—although implicit in the original problem—are not imposed. The LLM neglects these implicit constraints during formalization. Additionally, the Lean statement lacks the goal of proving the conditions under which the equality holds.

This geometric problem example clearly represents a failed formalization case, yet it still passed both the formal verification and consistency checks in the autoformalization pipeline. This indicates that LLMs have limited understanding of geometric problems. Although some prior work has attempted autoformalization of Euclidean geometry problems\cite{38}, even with improvements, the accuracy remains around $20\%$. Also the LeanEuclid dataset construction still relies heavily on manual annotation. Such formalization approaches are evidently unsuitable for building high-quality datasets. Therefore, this study temporarily excludes geometry problems from the original dataset.

The dataset also contains some mathematical problems where a single problem includes multiple subproblems. These subproblems may be unrelated, may share the same set of conditions, or may have a hierarchical relationship. Although the Lean system can handle multi-goal problems, for simplicity, multiple subproblems under the same problem number are split. Specifically, the original problem is divided into subproblems of the form “shared conditions + subgoal 1,” “shared conditions + subgoal 2,” and so forth. 

The tasks of filtering out geometry problems and splitting subgoals are also performed by the Deepseek-R1 model. After these processes, a total of $4,798$ natural language mathematical problems were retained.

\subsection{Dataset Construction and Evaluation}
The $4,798$ natural language mathematical problems were processed through the autoformalization pipeline in this study. After multiple sampling rounds and error feedback iterations, $4,481$ formalized statements passed formal verification, achieving a pass rate of $93.39\%$. Subsequently, after consistency check, a natural language–Lean aligned dataset comprising $3,922$ entries was constructed, corresponding to a formalization accuracy of $81.74\%$. Constructing this dataset cost $161,037,099$ tokens in total, including $22,492,280$ for prompting and $138,544,819$ for completion.

\begin{table}[t]
\caption{The result of formalization.}
\label{tab:stats}
\begin{center}
\begin{small}
\begin{sc}
\begin{tabular}{lrr}
        \toprule
        Class               & Number & Ratio \\
        \midrule
        Total                        & 4798          & 100\%         \\
        Formal verification          & 4481          & 93.39\%       \\
        \hspace{2em}Pass at one go   & 4287          & 89.35\%       \\
        \hspace{2em}Pass with error feedback & 194   & 4.04\%        \\
        Consistency check            & 3922          & 81.74\%       \\
        \hspace{2em}Pass at one go      & 3631        & 75.68\%       \\
        \hspace{2em}Pass with error feedback & 291    & 6.07\%        \\
        \bottomrule
    \end{tabular}
\end{sc}
\end{small}
\end{center}
\vspace{-0.9cm}
\end{table}

Among the 4,481 statements that passed formal verification, 4,287 were verified on the first attempt, while 194 were corrected and passed after error feedback, improving the formal verification pass rate by $4.04\%$. Of the $3,922$ entries that passed consistency check, $3,631$ passed initially, and $291$ were corrected and passed after error feedback, increasing the consistency check pass rate by $6.07\%$. Detailed data are presented in Table~\ref{tab:stats}. These results demonstrate that automated error feedback is highly effective in improving formalization accuracy.

\begin{figure}[ht]
\begin{center}
\centerline{\includegraphics[width=\columnwidth]{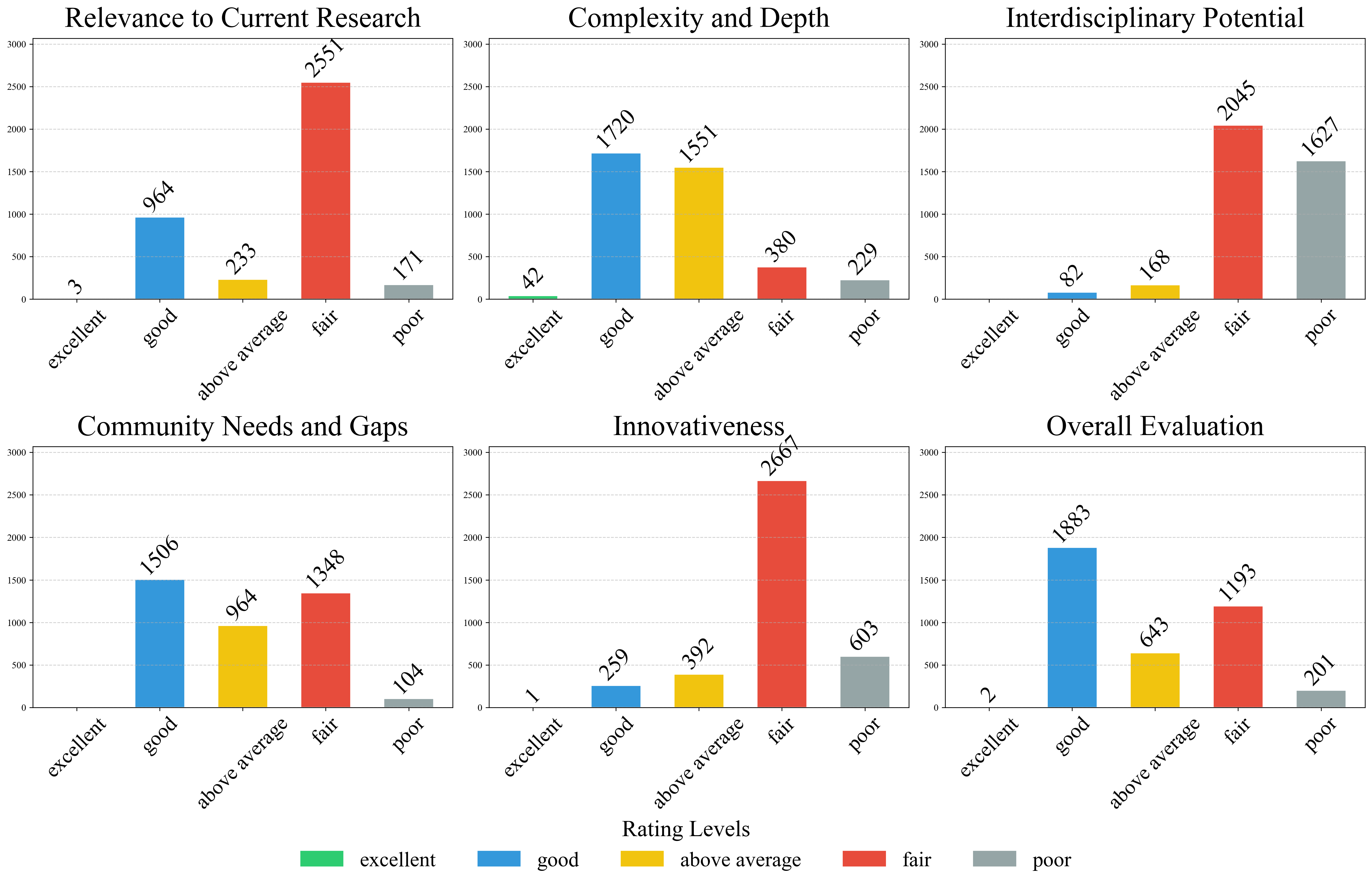}}
\vspace{-0.4cm}
\caption{Distribution of evaluation indicators.}
\label{fig:combined_metrics_eng}
\end{center}
\vspace{-1.2cm}
\end{figure}

This study further assesses dataset quality by evaluating each mathematical theorem along five dimensions using a large language model: relevance to current research, complexity and depth, interdisciplinary potential, community needs and gaps, and innovativeness. “Relevance to current research” examines whether the theorem addresses actively studied problems or concepts in mathematics or related fields; “complexity and depth” considers if the theorem sufficiently challenges existing theories and methods or offers significant insights or advances; “interdisciplinary potential” assesses opportunities for cross-disciplinary research, such as linking mathematics with computer science, physics, or biology; “community needs and gaps” evaluates whether the theorem fills identified demands or gaps within the Lean community or the broader mathematical community; and “innovativeness” gauges whether the theorem introduces new methods, concepts, or applications. Each dimension is rated as excellent, good, above average, fair, or poor, with an overall rating also provided. The results, shown in Figure~\ref{fig:combined_metrics_eng}, indicate strong performance in complexity and depth as well as community needs and gaps, but weaker performance in interdisciplinary potential and innovativeness. This pattern closely relates to the original data selection, as Olympiad-level problems tend to be relatively challenging, and the Lean community currently lacks data of comparable difficulty. Overall, $64.46\%$ of the data received ratings of above average or higher, suggesting a generally acceptable dataset quality.

\section{Experiments}
\subsection{Autoformalization Capability of Different LLMs}

This paper compares the formalization capabilities of Deepseek-R1, GPT-4o-mini, and Claude 3.7 Sonnet.
(1) Deepseek-R1\cite{44}: Released by the DeepSeek team in January 2025, it emphasizes high performance with low resource consumption. It supports complex logical deduction and long code generation, exhibiting “reflective” reasoning behavior, and demonstrates strong performance in mathematics, coding, and reasoning tasks.
(2) GPT-4o-mini\cite{45}: Released by OpenAI in July 2024, this smaller, more cost-effective model compared to GPT-4o excels in text and visual reasoning, mathematics, coding, and multimodal inference, featuring high resource efficiency and low inference latency.
(3) Claude 3.7 Sonnet\cite{10}: Released by Anthropic in February 2025, this model combines general large language model and reasoning capabilities, with self-reflective abilities during deliberation, and performs excellently in mathematics, physics, instruction following, coding, and many other tasks.

The selection of these three models primarily considers their strengths in reasoning and their low resource consumption. They are respectively applied as the formalization model, backtranslation model, and consistency check model within the autoformalization pipeline. Experiments were conducted on a random sample of 100 original problems (including geometry problems). The results, shown in Table~\ref{tab:model_compare}, indicate that at the cost of higher resource use and time, Deepseek-R1 significantly outperforms the other two models in formalization capability, achieving a final formalization accuracy of 43\%. Although the theorems formalized by GPT-4o-mini and Claude 3.7 Sonnet exhibit comparable pass rates in formal verification, the final number of theorems passing consistency check is substantially higher for Claude 3.7 Sonnet than for GPT-4o-mini.

\begin{table*}[t]
\vspace{-0.2cm}
\caption{Model performance comparison. The number of tokens and the time are both average values.}
\label{tab:model_compare}
\begin{center}
\begin{small}
\begin{sc}
\begin{tabular}{l *{2}{S[table-format=4.2] S[table-format=3.2] c}}
    \toprule
    \multirow{2}{*}{Model} & 
    \multicolumn{3}{c}{Formal verification} & 
    \multicolumn{3}{c}{consistency check} \\
    \cmidrule(lr){2-4} \cmidrule(lr){5-7}
    & {Token number} & {Time(s)} & {pass rate} & {Token number} & {Time(s)} & {pass rate} \\
    \midrule
    Deepseek-R1    & 10783.79 & 338.16 & 58\% & 3357.41 & 94.74 & 43\% \\
    GPT-4o-mini    & 1508.33 & 17.00 & 34\% & 737.20 &  3.97 & 11\% \\
    Claude 3.7 Sonnet     & 1721.87 & 43.28 & 31\% & 881.16 & 25.10 & 27\% \\
    \bottomrule
\end{tabular}
\end{sc}
\end{small}
\end{center}
\vspace{-0.4cm}
\end{table*}

\begin{table}[h]
\vspace{-0.2cm}
\caption{Model cross-validation results. The cell format is: \textit{Formal verification pass rate/consistency check pass rate}.}
\label{tab:cross-validation}
\begin{center}
\begin{sc}
\resizebox{\linewidth}{!}{
    \begin{tabular}{lccc}
        \toprule
        \multirow{2}{*}{formalization model} & \multicolumn{3}{c}{\textbf{consistency check model}} \\
        \cmidrule(lr){2-4}
        & Deepseek-R1 & GPT-4o-mini & Claude 3.7 Sonnet \\
        \midrule
        Deepseek-R1    & 58\% / 43\% & 58\% / 31\% & 58\% / 54\% \\
        GPT-4o-mini    & 34\% / 10\% & 34\% / 11\% & 34\% / 22\% \\
        Claude 3.7 Sonnet     & 31\% / 22\% & 31\% / 14\% & 31\% / 27\% \\
        \bottomrule
    \end{tabular}
}
\end{sc}
\end{center}
\vspace{-0.4cm}
\end{table}

To further investigate the formalization capabilities of the three models and their ability to assess semantic consistency, cross-experiments were conducted. Specifically, the same model was used for both formalization and back-translation, while a different model was employed for consistency checking. The experimental results are presented in Table~\ref{tab:cross-validation}. In the table, the first metric denotes the formal verification pass rate, and the second indicates the consistency check pass rate, representing the overall formalization accuracy.

It is observed that when using the same consistency checking model, Deepseek-R1 achieves significantly higher formal verification and consistency check pass rates. This indicates that Deepseek-R1 is better at ensuring both syntactic and semantic correctness during formalization. Although GPT-4o-mini’s formal verification pass rate is slightly higher than that of Claude 3.7 Sonnet, its consistency check pass rate is notably lower. This suggests that while GPT-4o-mini adheres to Lean’s syntax rules, it struggles to accurately capture the original mathematical problem’s intent.

\begin{table}[h]
\vspace{-0.5cm}
\caption{Evaluation matrix for consistency checks of different models. Experiments were based on formalization model Deepseek-R1.}
\vspace{-0.3cm}
\label{tab:model-metrics}
\begin{center}
\resizebox{\linewidth}{!}{
\begin{sc}
    \begin{tabular}{lcccc}
    \toprule
    model name & Accuracy & Precision & Recall & F1 score \\
    \midrule
    Deepseek-R1   & 74.1\% & 69.8\% & 93.8\% & 80.0\% \\
    GPT-4o-mini   & 74.1\% & 77.4\% & 75.0\% & 76.2\% \\
    Claude-3-7    & 58.6\% & 57.4\% & 96.9\% & 72.1\% \\
    \bottomrule
    \end{tabular}
\end{sc}
}
\end{center}
\vspace{-0.5cm}
\end{table}

Furthermore, when the same model is used for both formalization and backtranslation, Claude 3.7 Sonnet achieves the highest consistency check pass rate when employed as the consistency checker, while the results for Deepseek-R1 and GPT-4o-mini fluctuate slightly. Manual analysis of the formalized mathematical problems shows that Claude 3.7 Sonnet tends to give affirmative answers during consistency checks, meaning it more often classifies mathematically mismatched natural language–Lean pairs as correct formalizations. In contrast, GPT-4o-mini adopts a stricter consistency criterion, labeling more mathematically matched pairs as incorrect formalizations. Compared to these, Deepseek-R1 delivers the best overall performance, balancing precision and recall effectively; detailed metrics are provided in Table~\ref{tab:model-metrics}. One notable exception is that when GPT-4o-mini is used for both formalization and back-translation, its consistency check pass rate exceeds that of Deepseek-R1—likely due to an inflated accuracy from using the same model for both formalization and consistency checking.

\subsection{Effect of Few-shot Learning}

The experiments were conducted on a subset of $57$ data points after excluding geometric problems; the following two experiments also used this subset. Table~\ref{tab:few_shot} presents the results of few-shot learning experiments. These experiments involved error feedback and five rounds of sampling. Few-shot learning was applied only during the first formalization attempt of each natural language mathematical theorem; subsequent prompts contained only error feedback without translation examples. The results indicate that few-shot learning indeed increased the one-shot formal verification pass rate. However, it remains unclear whether this contributed to differences in overall formalization accuracy. Although the overall accuracy difference reached as high as 8.76\%, this increase mainly arose from the second formalization of theorems that initially failed consistency checks. At the time of the first consistency check, formalization accuracies were comparable (44\% vs. 45\%). During second formalization, only error feedback was provided without correct sample prompts. Whether the formal statements that passed verification but failed consistency checks also served as few-shot learning data and contributed to accuracy improvement under few-shot learning warrants further study. Nonetheless, the data show that few-shot learning overall improved formalization accuracy.

\begin{table}[h]
\vspace{-0.3cm}
\caption{The effect of few-shot learning on formalization accuracy.}
\label{tab:few_shot}
\begin{center}
\resizebox{\linewidth}{!}{
\begin{sc}
    \begin{tabular}{cccccc c}
    \toprule
    \multirow{2}{*}{class} & 
    \multicolumn{2}{c}{formal verification} & 
    \multicolumn{2}{c}{consistency check} & 
    \multirow{2}{*}{pass rate} \\
    \cmidrule(lr){2-3} \cmidrule(lr){4-5}
    & first pass & second pass & first pass & second pass & \\ 
    \midrule
    few-shot & 51 & 4 & 44 & 8 & 91.22\% \\ 
    zero-shot     & 45 & 5 & 45 & 2 & 82.46\% \\ 
    \bottomrule
    \end{tabular}
\end{sc}
}
\end{center}
\vspace{-0.3cm}
\end{table}

\subsection{Effect of Error Feedback}

Table~\ref{tab:error_info} demonstrates the improvement in formalization accuracy achieved through error feedback. To ensure that formalization results were not influenced by the number of sampling attempts, experiments without error feedback also retained the second translation attempt for theorems that failed the first formal verification or consistency check, but no error information was provided. Analysis of the results shows that error messages returned by the formal verification had a limited effect on improving the formal verification pass rate; the number of second-time passes with error feedback (4) was roughly the same as that without error feedback (3). This may be because the error messages from formal verification are structured data that the translation model finds difficult to interpret, or because the initial formal verification pass rate was already high and those failed formal verification were ill-defined problems. In contrast, for the consistency check, error feedback significantly increased the pass rate on the second attempt, thereby improving the overall formalization accuracy by $10.52\%$. This confirms the positive impact of error feedback on enhancing formalization accuracy.

\begin{table}[h]
\vspace{-0.3cm}
\caption{The effect of error feedback on formalization accuracy.}
\label{tab:error_info}
\begin{center}
\resizebox{\linewidth}{!}{
\begin{sc}
    \begin{tabular}{cccccc c}
    \toprule
    \multirow{2}{*}{class} & 
    \multicolumn{2}{c}{formal verification} & 
    \multicolumn{2}{c}{consistency check} & 
    \multirow{2}{*}{pass rate} \\
    \cmidrule(lr){2-3} \cmidrule(lr){4-5}
    & first pass & second pass & first pass & second pass & \\ 
    \midrule
    with error feedback & 51 & 4 & 44 & 8 & 91.22\% \\ 
    no error feedback   & 52 & 3 & 46 & 0 & 80.70\% \\ 
    \bottomrule
    \end{tabular}
\end{sc}
}
\end{center}
\vspace{-0.3cm}
\end{table}

\subsection{Effect of Increasing Sampling Number}

Table~\ref{tab:sample} presents experiments on increasing the sampling number, where the sampling number refers to how many times the same statement is input into the autoformalization pipeline. Experimental data show that multiple samplings significantly improve the pass rates of both the first formal verification and consistency check, resulting in an overall increase of $31.57\%$ in formalization accuracy. This performance boost is mainly attributed to the randomness and diversity of outputs generated by large language models.

\begin{table}[h]
\vspace{-0.3cm}
\caption{The effect of increasing sampling number on formalization accuracy.}
\vspace{-0.3cm}
\label{tab:sample}
\begin{center}
\resizebox{\linewidth}{!}{
\begin{sc}
    \begin{tabular}{cccccc c}
    \toprule
    \multirow{2}{*}{class} & 
    \multicolumn{2}{c}{formal verification} & 
    \multicolumn{2}{c}{consistency check} & 
    \multirow{2}{*}{pass rate} \\
    \cmidrule(lr){2-3} \cmidrule(lr){4-5}
    & first pass & second pass & first pass & second pass & \\ 
    \midrule
    sample @1 & 36 & 5 & 28 & 6 & 59.65\% \\ 
    sample @5   & 51 & 4 & 44 & 8 & 91.22\% \\ 
    \bottomrule
    \end{tabular}
\end{sc}
}
\end{center}
\vspace{-0.3cm}
\end{table}

\subsection{Testing as a Benchmark for Automated Theorem Provers}

To assess the relative difficulty of different formal mathematical datasets, three state-of-the-art automated theorem provers -- Kimina-Prover, Goedel-Prover, and DeepSeek-Prover-V1.5-RL -- are benchmarked against established datasets MiniF2F, ProofNet, and FormalMATH\cite{54} alongside this newly constructed formal language dataset.

Each verification task was evaluated over $32$ runs on $1,000$ randomly sampled problems from $9,787$ formal statements to improve the statistical reliability of formal validation, with consistent hyperparameters across all provers. As shown in Table~\ref{tab:as_benchmark}, our dataset achieved intermediate pass rates of $15.7\%$ (Goedel-Prover) and $13.0\%$ (DeepSeek-Prover), positioning its difficulty between ProofNet ($15.2\%-16.0\%$) and FormalMATH ($10.2\%-16.5\%$). These results demonstrate that our dataset not only presents a meaningful level of challenge but also effectively differentiates the performance of provers on moderately complex formal reasoning tasks, highlighting its competitiveness as a benchmark.

\begin{table}[h]
\vspace{-0.3cm}
\caption{Test results of different automated theorem provers. Each verification task was evaluated over $32$ runs on $1,000$ randomly sampled formal problems.}
\label{tab:as_benchmark}
\begin{center}
\resizebox{\linewidth}{!}{
\begin{sc}
    \begin{tabular}{lccc}
    \toprule
    dataset & Kimina-Prover & Goedel-Prover & DeepSeek-Prover-V1.5-RL \\
    \midrule
    MiniF2F          & 63.1\% & 57.6\% & 50.0\% \\
    ProofNet         & - & 15.2\% & 16.0\% \\
    FormalMATH       & 16.5\% & 13.5\% & 10.2\% \\
    \dataset         & 16.4\% & 15.7\% & 13.0\% \\
    \bottomrule
    \end{tabular}
\end{sc}
}
\end{center}
\vspace{-0.3cm}
\end{table}

\section{Conclusion}

This paper reviews existing natural language–Lean aligned mathematical problem datasets and points out that datasets at competition-level difficulty tend to be relatively small, while other automatically constructed formalization datasets often lack guaranteed difficulty. To address this, we propose an autoformalization pipeline with error feedback inspired by the Lean Workbook formalization framework. Unlike the Lean Workbook pipeline, which relies on expert annotations and iterative refinement, our pipeline can automatically generate error feedback and perform iterative prompt optimization based on this feedback. Moreover, all models used in this pipeline are off-the-shelf general-purpose large language models, further demonstrating the reasoning and formalization capabilities of large language models while greatly reducing deployment costs.
Using this autoformalization pipeline, we constructed a natural language–Lean aligned dataset with a formalization semantic accuracy of $81.74\%$, ultimately containing $3,922$ mathematical problems in natural language and $9,787$ in Lean at the difficulty level of the Olympiad mathematics competitions. Our dataset strikes a balances bwtween scale and difficulty. According to quality assessments, $64.46\%$ of the formalized data received evaluations of at least “above average.”
Furthermore, a series of ablation experiments confirmed the formalization advantages of the Deepseek-R1 model over GPT-4o-mini and Claude 3.7 Sonnet, and verified the positive effects of few-shot learning, error feedback, and increased per-problem sampling on improving formalization accuracy. We evaluated three state-of-the-art automated theorem provers on a subset of the \dataset\ dataset, highlighting its challenging nature and its value as a benchmark for formal reasoning tasks.

Regarding the raw data, due to the expressive limitations of the Lean language, this study excluded all geometric mathematical problems. Geometric problems hold a crucial position not only in the field of competitions but also across the entire domain of mathematical research. Therefore, further development of Lean in this area, or the emergence of new formal languages with better support for geometric mathematics, is highly anticipated.
In terms of autoformalization, due to constraints on computational resources and time, this study only performed 5 samplings per problem. Under more favorable conditions, increasing the number of samplings could potentially yield higher formalization accuracy. Additionally, manual inspection of the dataset revealed that even when using the most capable Deepseek-R1 model, misjudgments still occur during consistency checks. This indicates that the currently most advanced general-purpose large language models still have limitations in formalization and consistency verification, thereby demonstrating the inherent challenges associated with autoformalization.




\section*{Impact Statement}

This paper presents work whose goal is to advance the field of 
Machine Learning. There are many potential societal consequences 
of our work, none which we feel must be specifically highlighted here.

\section*{Acknowledgments}
This paper is partially supported by grants from the National Key Research and Development Program of China with Grant No. 2023YFC3341203 and the National Natural Science Foundation of China
(NSFC Grant Number 62276002). 


\bibliography{example_paper}
\bibliographystyle{icml2025}

\newpage
\appendix
\onecolumn
\section{Case study}

\label{sec:formalization-challenges}

In the process of translating mathematical problems into formal statements acceptable to theorem provers, we have identified three main challenges: semantic gaps in problem representation, goal definition issues, and missing conditions. This section analyzes these challenges through concrete case studies.

\subsection{Semantic Gaps in Problem Representation}
\label{subsec:semantic-gap}

Semantic gaps occur when formalized statements fail to accurately capture the mathematical essence of original problems. These issues are particularly prominent in combinatorial mathematics and recursive structure problems.

\begin{tcolorbox}[
    colback=white,      
    colframe=black,     
    arc=0pt,            
    boxrule=1pt,        
    width=0.9\linewidth,
    center,             
    left=5mm,           
    right=5mm,          
    top=3mm,            
    bottom=3mm,         
    breakable           
]
\textbf{Original problem}: $n$ children sit in a circle. A teacher distributes candies clockwise starting from one child: gives to first, skips one, gives to next, skips two, gives to next, and so on (increasing skips). Find $n$ such that eventually every child receives at least one candy.

\medskip

\textbf{Formalized output}:

\begin{lstlisting}[language=lean,basicstyle=\ttfamily,frame=none]
theorem candy_distribution (n : ℕ) : 
  ∃ m : ℕ, ∀ i : Fin n, ∃ k : ℕ, 
  k < m ∧ (k * (k + 3) / 2) % n = i := by sorry
\end{lstlisting}

\end{tcolorbox}

\textbf{Analysis}:
The formalization exhibits three fundamental discrepancies. Primarily, there exists a critical mismatch in the core mathematical expression: the adopted quadratic form $(k(k+3)/2)$ fails to correspond with the actual skipping pattern, where triangular numbers $k(k+1)/2$ would be the correct characterization of the candy distribution rule. More critically, the conclusion remains incomplete as the formalization merely proves the existence of qualifying $n$ without characterizing its essential property (specifically that $n$ should be a power of two, $n=2^k$), thereby depriving the result of practical applicability.

\begin{tcolorbox}[
    colback=white,      
    colframe=black,     
    arc=0pt,            
    boxrule=1pt,        
    width=0.9\linewidth,
    center,             
    left=5mm,           
    right=5mm,          
    top=3mm,            
    bottom=3mm,         
    breakable           
]

\textbf{Original problem}: Given triangle with sides $a,b,c$, construct new triangle with sides $s-a,s-b,s-c$ where $s$ is semiperimeter. Repeat until construction impossible. Find initial triangles allowing infinite iteration.

\medskip
\medskip

\textbf{Formalized output}:

\begin{lstlisting}[language=lean,basicstyle=\ttfamily,frame=none]
theorem indefinite_semiperimeter_triangle (a b c : ℝ) :
  (∀ s : ℝ, s = (a + b + c) / 2 → 
    ∃ a' b' c' : ℝ, 
      a' = s - a ∧ b' = s - b ∧ c' = s - c ∧
      a' + b' > c' ∧ b' + c' > a' ∧ c' + a' > b' ∧ 
      ∀ n : ℕ, 
        ∃ a'' b'' c'' : ℝ,
          a'' = (a' + b' + c') / 2 - a' ∧ 
          b'' = (a' + b' + c') / 2 - b' ∧ 
          c'' = (a' + b' + c') / 2 - c' ∧
          a'' + b'' > c'' ∧ 
          b'' + c'' > a'' ∧ 
          c'' + a'' > b'') → 
  a = b ∧ b = c := by sorry
\end{lstlisting}

\end{tcolorbox}

\textbf{Analysis}:
The formalization contains three critical flaws in recursive construction. Primarily, it demonstrates insufficient recursion depth by explicitly building only two iteration levels (from $(a,b,c)$ to $(a',b',c')$ to $(a'',b'',c'')$), failing to establish a genuine infinite recursion structure. Secondly, there exists structural discontinuity between iteration levels - the inputs of the second iteration $(a',b',c')$ lack proper mathematical connection to the outputs of the first iteration $(a,b,c)$, creating an open-loop system that breaks recursive coherence. 

\subsection{Goal Definition Issues}
\label{subsec:goal-definition}

Solving-oriented problems ("Find" type) often suffer from missing or misplaced goals, especially when requiring specific values or function expressions.

\begin{tcolorbox}[
    colback=white,      
    colframe=black,     
    arc=0pt,            
    boxrule=1pt,        
    width=0.9\linewidth,
    center,             
    left=5mm,           
    right=5mm,          
    top=3mm,            
    bottom=3mm,         
    breakable           
]
\textbf{Original problem}: Find all functions $f:\mathbb{R}\to\mathbb{R}$ satisfying:
(i) $f(x+y) \in [f(x)+f(y), f(x)+f(y)+1]$
(ii) $x\in[0,1) \Rightarrow f(0) \geq f(x)$
(iii) $f(1)=1 \land -f(-1)=1$

\medskip
\medskip

\textbf{Formalized output}:
\begin{lstlisting}[language=lean,basicstyle=\ttfamily,frame=none]
theorem translated_theorem :
  ∃ f : ℝ → ℝ, 
  (∀ x y : ℝ, f x + f y + 1 ≥ f (x + y) ∧ f (x + y) ≥ f x + f y) ∧ 
  (∀ x : ℝ, 0 ≤ x ∧ x < 1 → f 0 ≥ f x) 
  ∧ f 1 = 1 ∧ -f (-1) = 1 := by sorry
\end{lstlisting}

\end{tcolorbox}

\textbf{Analysis}:
The primary issue is goal misalignment: while the original problem demands "finding all satisfying functions", the formalized version merely proves the existence of such functions and omits the concrete solution set, representing a substantial deviation from the original requirement. 

\begin{tcolorbox}[
    colback=white,      
    colframe=black,     
    arc=0pt,            
    boxrule=1pt,        
    width=0.9\linewidth,
    center,             
    left=5mm,           
    right=5mm,          
    top=3mm,            
    bottom=3mm,         
    breakable           
]
\textbf{Original problem}: Let $\{f(n)\}$ be strictly increasing sequence of positive integers where the $n$-th missing positive integer is $f(f(n))+1$. Find $f(240)$.

\medskip
\medskip

\textbf{Formalized output}:
\begin{lstlisting}[language=lean,basicstyle=\ttfamily,frame=none]
theorem test
  (f : ℕ → ℕ)
  (h₀ : ∀ n, 0 < f n)
  (h₁ : ∀ n m, n < m → f n < f m)
  (h₂ : ∀ n, ∃ m, m = f (f n) + 1 ∧ ∀ k < m, (∃ i, f i = k) → ∃ i ≤ k, f i = k) :
  f 240 = 38808 := by sorry
\end{lstlisting}

\end{tcolorbox}

\textbf{Analysis}:
The formalization exhibits two fundamental flaws. Firstly, it fails to properly characterize the "n-th missing positive integer" - while it expresses the relationship $m=f(f(n))+1$ using existential quantification, it lacks the necessary cardinality constraints to ensure this indeed represents the n-th (rather than any arbitrary) missing number. More critically, the formalized conclusion's assertion that $f(240)=38808$ is incorrect, revealing a serious validation failure. 

\subsection{Missing Conditions}
\label{subsec:missing-conditions}

Omission of critical conditions during formalization can lead to incorrect conclusions, particularly dangerous in number theory problems.

\begin{tcolorbox}[
    colback=white,      
    colframe=black,     
    arc=0pt,            
    boxrule=1pt,        
    width=0.9\linewidth,
    center,             
    left=5mm,           
    right=5mm,          
    top=3mm,            
    bottom=3mm,         
    breakable           
]
\textbf{Original problem}: Let $k\geq 14$, $p_k$ the largest prime strictly less than $k$ (given $p_k\geq 3k/4$). Let $n$ be composite. Prove:
(a) If $n=2p_k$, then $n\nmid (n-k)!$
(b) If $n>2p_k$, then $n\mid (n-k)!$

\medskip
\medskip

\textbf{Formalized output}:
\begin{lstlisting}[language=lean,basicstyle=\ttfamily,frame=none]
theorem part_a (k : ℕ) (hk : k ≥ 14)
  (p_k : ℕ) (hp_k : Nat.Prime p_k)
  (h : p_k < k) (h₁ : p_k ≥ 3*k/4)
  (n : ℕ) (hn : ¬Nat.Prime n)
  (hn_eq : n = 2 * p_k) : 
  ¬n | (n - k).factorial := by sorry
\end{lstlisting}
\end{tcolorbox}

\textbf{Analysis}:
The formalization exhibits several critical limitations. Most notably, it fails to explicitly specify that $p_k$ must be the largest prime strictly less than $k$. While the theorem states that $p_k$ is prime and satisfies $p_k < k$, the absence of the maximality condition introduces a fundamental flaw—the conclusion may become invalid if multiple primes satisfy $p_k < k$. 

\section{Comparison with Other Datasets}
To validate the reliability of our dataset, we selected some mathematical problems which overlap with FMC from FIMO\cite{53}, CombiBench\cite{52}, and FormalMATH\cite{54}, and compared their formalization accuracy and stylistic differences. All three datasets adopt Lean as their formal language.

\subsection{FIMO}
The FIMO\cite{53} dataset contains problems drawn from the International Mathematical Olympiad (IMO) Shortlisted Problems. Below is an example that appears in both FMC and FIMO.

\begin{tcolorbox}[
    colback=white,      
    colframe=black,     
    arc=0pt,            
    boxrule=1pt,        
    width=0.9\linewidth,
    center,             
    left=5mm,           
    right=5mm,          
    top=3mm,            
    bottom=3mm,         
    breakable           
]
\textbf{Original problem}: Prove that
$$
\frac{x^{2}}{(x-1)^{2}}+\frac{y^{2}}{(y-1)^{2}}+\frac{z^{2}}{(z-1)^{2}} \geq 1
$$
for all real numbers $x, y, z$, each different from 1, and satisfying $xyz=1$.

\medskip
\medskip

\textbf{FMC}:
\begin{lstlisting}[language=lean,basicstyle=\ttfamily,frame=none]
theorem test
  (x y z : ℝ)
  (h₀ : x ≠ 1)
  (h₁ : y ≠ 1)
  (h₂ : z ≠ 1)
  (h₃ : x * y * z = 1) :
  x^2 / (x - 1)^2 + y^2 / (y - 1)^2 + z^2 / (z - 1)^2 ≥ 1 := by sorry
\end{lstlisting}

\medskip
\medskip

\textbf{FIMO}:
\begin{lstlisting}[language=lean,basicstyle=\ttfamily,frame=none]
theorem fimo_2008_algebra_p2_1
  (x y z : ℝ)
  (h₀ : x ≠ 1 ∧ y ≠ 1 ∧ z ≠ 1)
  (h₁ : x * y * z = 1) :
  x^2 / (x - 1)^2 + y^2 / (y - 1)^2 + z^2 / (z - 1)^2 ≥ 1 :=
begin
  sorry
end
\end{lstlisting}
\end{tcolorbox}

This problem involves simple conditions and a clear goal so that both formal statements accurately capture the intended meaning with only minor differences. Specifically, FMC lists the assumptions 
$x \neq 1, y \neq 1$ and $z \neq 1$ separately, while FIMO combines them into a single conjunctive premise. From a formal reasoning perspective, FMC's formulation facilitates usage of assumptions, whereas FIMO's version offers improved readability. Additionally, the two statements' placeholder styles differ: \texttt{by sorry} versus \texttt{begin...sorry...end}, reflecting style discrepancy between Lean versions.

\subsection{FormalMATH}
The FormalMATH\cite{54} dataset spans a wide range of topics, from high school Olympiad problems to undergraduate-level theorems. Below are two examples found in both FMC and FormalMATH.

\begin{itemize}
    \item Case 1
\end{itemize}

\begin{tcolorbox}[
    colback=white,      
    colframe=black,     
    arc=0pt,            
    boxrule=1pt,        
    width=0.9\linewidth,
    center,             
    left=5mm,           
    right=5mm,          
    top=3mm,            
    bottom=3mm,         
    breakable           
]
\textbf{Original problem}: Find all functions \( f : \mathbb{R} \to \mathbb{R} \) such that for all real numbers \( x, y \),
\[
f(f(x) + y) = f(x^2 - y) + 4y f(x).
\]

\medskip
\medskip

\textbf{FMC}:
\begin{lstlisting}[language=lean,basicstyle=\ttfamily,frame=none]
theorem test
  (f : ℝ → ℝ)
  (h₀ : ∀ x y, f (f x + y) = f (x ^ 2 - y) + 4 * y * f x) :
  f = 0 ∨ f = fun x => x ^ 2 := by sorry
\end{lstlisting}

\medskip
\medskip

\textbf{FormalMATH}:
\begin{lstlisting}[language=lean,basicstyle=\ttfamily,frame=none]
theorem olymid_ref_base_11031 (f : ℝ → ℝ) : 
  (∀ x y, f (f x + y) = f (x ^ 2 - y) + 4 * y * f x) ↔
  ∀ x, f x = 0 ∨ f x = x ^ 2 := by
\end{lstlisting}
\end{tcolorbox}

Similar to the FIMO example, both formalizations are largely consistent and accurate with respect to the natural language description.

\begin{itemize}
    \item Case 2
\end{itemize}

\begin{tcolorbox}[
    colback=white,      
    colframe=black,     
    arc=0pt,            
    boxrule=1pt,        
    width=0.9\linewidth,
    center,             
    left=5mm,           
    right=5mm,          
    top=3mm,            
    bottom=3mm,         
    breakable           
]
\textbf{Original problem}: Show that for any integer $n \geq 2$ the sum of the fractions $\frac{1}{a b}$, where $a$ and $b$ are relatively prime positive integers such that $a<b \leq n$ and $a+b>n$, equals $\frac{1}{2}$.
(Integers $a$ and $b$ are called relatively prime if the greatest common divisor of $a$ and $b$ is 1.)

\medskip
\medskip

\textbf{FMC}:
\begin{lstlisting}[language=lean,basicstyle=\ttfamily,frame=none]
theorem test
  (n : ℕ)
  (h₀ : 2 ≤ n) :
  Finset.sum (Finset.filter (\lambda ab : ℕ × ℕ => ab.1 < ab.2 ∧ ab.2 ≤ n ∧ ab.1 + ab.2 > n ∧ Nat.gcd ab.1 ab.2 = 1) (Finset.product (Finset.Icc 1 n) (Finset.Icc 1 n))) (\lambda ab => 1 / (ab.1 * ab.2 : ℚ)) = 1 / 2 := by sorry
\end{lstlisting}

\medskip
\medskip

\textbf{FormalMATH}:
\begin{lstlisting}[language=lean,basicstyle=\ttfamily,frame=none]
theorem olymid_ref_base_11032 {n : ℕ} (hn : 2 ≤ n) :
  ∑' a : ℕ, ∑' b : ℕ,
  (if (a < b ∧ b ≤ n ∧ Nat.Coprime a b ∧ a + b > n) then (1 / ((a * b) : ℚ)) else 0) =
  1 / 2 := by
\end{lstlisting}
\end{tcolorbox}

From a mathematical perspective, both versions express the same content rigorously. The key difference lies in stylistic preference: FMC uses \texttt{Finset} to denote finite sets in Lean, aligning with Lean’s programming idioms, while FormalMATH opts for summation notation and \texttt{if...then} logic, which more closely resembles traditional mathematical expressions and improves human readability.

\subsection{CombiBench}
The CombiBench\cite{52} dataset contains 100 carefully selected combinatorial problems formalized by mathematical experts, including all combinatorial IMO problems since 2000. Below is an overlapping example found in both FMC and CombiBench.

\begin{tcolorbox}[
    colback=white,      
    colframe=black,     
    arc=0pt,            
    boxrule=1pt,        
    width=0.9\linewidth,
    center,             
    left=5mm,           
    right=5mm,          
    top=3mm,            
    bottom=3mm,         
    breakable           
]
\textbf{Original problem}: A magician has one hundred cards numbered 1 to 100 . He puts them into three boxes, a red one, a white one, and a blue one, so that each box contains at least one card. A member of the audience draws two cards from two different boxes and announces the sum of numbers on those cards. Given this information, the magician locates the box from which no card has been drawn. How many ways are there to put the cards in the three boxes so that the trick works? (Two ways are considered different if at least one card is put into a different box.)

\medskip
\medskip

\textbf{FMC}:
\begin{lstlisting}[language=lean,basicstyle=\ttfamily,frame=none]
theorem test :
  let cards : Finset ℕ := Finset.range 100
  let boxes : Finset (Finset ℕ) := {∅, ∅, ∅} -- Placeholder for actual box definitions
  Finset.card (Finset.filter (fun b : Finset ℕ => Finset.Nonempty b) boxes) = 3 ∧
  (∀ b₁ b₂ : Finset ℕ, b₁ ∈ boxes → b₂ ∈ boxes → b₁ ≠ b₂ → ∀ x y : ℕ, x ∈ b₁ → y ∈ b₂ → 
  ∀ b₃ b₄ : Finset ℕ, b₃ ∈ boxes → b₄ ∈ boxes → b₃ ≠ b₄ → ∀ z w : ℕ, z ∈ b₃ → w ∈ b₄ → 
  x + y ≠ z + w) := by sorry
\end{lstlisting}

\medskip
\medskip

\textbf{CombiBench}:
\begin{lstlisting}[language=lean,basicstyle=\ttfamily,frame=none]
abbrev Cards := Finset.Icc 1 100
abbrev Boxes := Fin 3
abbrev Trick := ℕ → Boxes

def trick_works (f : Cards → Boxes) (t : Trick) : Prop :=
  ∀ c₁ c₂ : Cards,
  -- given the sum of two cards from box 0 and box 1 then the trick gives the result of box 2
  (f c₁ = 0 → f c₂ = 1 → t (c₁.1 + c₂.1) = 2) ∧
  -- given the sum of two cards from box 0 and box 2 then the trick gives the result of box 1
  (f c₁ = 0 → f c₂ = 2 → t (c₁.1 + c₂.1) = 1) ∧
  -- given the sum of two cards from box 1 and box 2 then the trick gives the result of box 0
  (f c₁ = 1 → f c₂ = 2 → t (c₁.1 + c₂.1) = 0)

theorem imo_2000_p4 (good_allocations : Finset (Cards → Boxes))
    (h : ∀ f, f ∈ good_allocations ↔ Function.Surjective f ∧ ∃ (t : Trick), trick_works f t) :
    good_allocations.card = imo_2000_p4_solution := by sorry
\end{lstlisting}
\end{tcolorbox}

In this case, the formal statement in FMC does not precisely align with the original natural language description. The problem specifies that, given the sum of two cards drawn from two different boxes, the magician can uniquely identify the box from which no card was drawn. This implies that the set of such pairwise sums for each pair of boxes must be disjoint, effectively serving as a signature for the third box.

However, the FMC version fails to explicitly ensure that $(b_1, b_2)$ and $(b_3, b_4)$ refer to two distinct pairs of boxes, violating the original problem constraints. Moreover, the formal statement does not specify the desired conclusion (i.e., the number of valid configurations), resulting in a missing goal. The approach also relies on explicit enumeration rather than set-based abstractions, making the expression unnecessarily verbose.

In contrast, the CombiBench version resolves these issues by introducing explicit box indices and counting mechanisms, leading to a clearer and more faithful formalization. This comparison suggests that, for combinatorial problems, LLM-generated formalizations still fall short, and statements formalized by mathematical experts retain a significant advantage.

\section{Prompts}
\subsection{Formal Translation Prompts}
There are two different stages of formal translation, the first goes with few-shot prompting and the second goes with error feedback. Following is the prompt for translation with few-shot learning.

\begin{tcolorbox}[
    colback=white,      
    colframe=black,     
    arc=0pt,            
    boxrule=1pt,        
    width=0.9\linewidth,
    center,             
    left=5mm,           
    right=5mm,          
    top=3mm,            
    bottom=3mm,         
    breakable           
]
    A math theorem in natural language will be provided and please translate it into a Lean4 theorem. 
    Please only return the translation (Lean4 code) and no analysis, no mathlib4 import, no comments, no proof, no reasoning. 
    Use ":= by sorry" as a placeholder for proof. Here are some examples for it: \{few\_shot\}. Following the examples above, 
    translate the next problem into Lean4: \{problem\}
\end{tcolorbox}

Following is the prompt for translation with error feedback.

\begin{tcolorbox}[
    colback=white,      
    colframe=black,     
    arc=0pt,            
    boxrule=1pt,        
    width=0.9\linewidth,
    center,             
    left=5mm,           
    right=5mm,          
    top=3mm,            
    bottom=3mm,         
    breakable           
]
    A math theorem in natural language will be provided and please translate it into a Lean4 theorem. 
    Please only return the translation (Lean4 code) and no analysis, no mathlib4 import, no comments, no proof, no reasoning. 
    Use ":= by sorry" as a placeholder for proof. Here is the theorem in natural language: \{problem\}. Before your translation, 
    note that this problem has been mistranslated as the following. Concrete errors have been listed and please avoid similar 
    mistakes when translating it again. Mistranslation: \{failed\_info\}
\end{tcolorbox}

\subsection{Backtranslation Prompts}
Following is the prompt for backtranslation.

\begin{tcolorbox}[
    colback=white,      
    colframe=black,     
    arc=0pt,            
    boxrule=1pt,        
    width=0.9\linewidth,
    center,             
    left=5mm,           
    right=5mm,          
    top=3mm,            
    bottom=3mm,         
    breakable           
]
    \textnormal{Convert the formal statement into natural language:\textbackslash n\texttt{\`{}\`{}\`{}}lean\textbackslash n\{prompt\}\textbackslash n\texttt{\`{}\`{}\`{}} \\
    Please only return the translation and no analysis.
    }
\end{tcolorbox}

\subsection{Consistency Check Prompts}
Following is the prompt for consistency check.

\begin{tcolorbox}[
    colback=white,      
    colframe=black,     
    arc=0pt,            
    boxrule=1pt,        
    width=0.9\linewidth,
    center,             
    left=5mm,           
    right=5mm,          
    top=3mm,            
    bottom=3mm,         
    breakable           
]
\textnormal{Please check whether the following two math problems is same or different in their mathematical essence: \\
Problem 1: \{origin\} \\
Problem 2: \{back\} \\
Please consider each statement in two problems, they are different if any condition or any goal is different. Return in the following format: \\
'''\{"Same": true/false, "Analysis": "Summarize their consistency and difference in brief"\}'''
}
\end{tcolorbox}

\subsection{Dataset Evaluation Prompts}
Our rating prompt is adapted from the paper \textit{DeepSeek-Prover: Advancing Theorem Proving in LLMs through Large-Scale Synthetic Data}. Following is the prompt.

\begin{tcolorbox}[
    colback=white,
    colframe=black,
    arc=0pt,
    boxrule=1pt,
    width=0.9\linewidth,
    center,
    left=5mm,
    right=5mm,
    top=3mm,
    bottom=3mm,
    breakable
]

To evaluate whether a formal Lean4 statement will be of interest to the community:  \textbackslash n\texttt{```lean}\textbackslash n\{\texttt{prompt}\}\textbackslash n\texttt{```} \\
\vspace{1ex}

Please consider the following criteria:

1. \textbf{Relevance to Current Research}: Does the statement address a problem or concept that is actively being researched in mathematics or related fields? Higher relevance scores indicate greater potential interest. \\

2. \textbf{Complexity and Depth}: Is the statement complex enough to challenge existing theories and methodologies, yet deep enough to provide significant insights or advancements? Complexity and depth showcase Lean4's capabilities and attract interest. \\

3. \textbf{Interdisciplinary Potential}: Does the statement offer opportunities for interdisciplinary research, connecting mathematics with other fields such as computer science, physics, or biology? Interdisciplinary projects often garner wide interest. \\

4. \textbf{Community Needs and Gaps}: Does the statement fill an identified need or gap within the Lean4 community or the broader mathematical community? Addressing these needs directly correlates with interest. \\

5. \textbf{Innovativeness}: How innovative is the statement? Does it propose new methods, concepts, or applications? Innovation drives interest and engagement. \\

Customize your evaluation for each problem accordingly, assessing it as \texttt{'excellent'}, \texttt{'good'}, \texttt{'above average'}, \texttt{'fair'} or \texttt{'poor'}. \\

You should respond in the following JSON format for each statement: \\

\texttt{\{%
 "Analysis ": (Provide a brief justification for each score, highlighting why the statement scored as it did across the criteria. Rate the statement as 'excellent', 'good', 'above average', 'fair' or 'poor' for all aspects respectively. Format \{%
 "Relevance to Current Research ": \{%
 "rating ": (rating),  "reason ": (reason)\},%
 "Complexity and Depth ": \{%
 "rating ": (rating),  "reason ": (reason)\}, ...\}), %
 "Assessment ": (Based on the criteria, rate the statement as 'excellent', 'good', 'above average', 'fair' or 'poor'.)%
\}}
\end{tcolorbox}

\end{document}